\documentclass{llncs}
 
\usepackage{amsmath} 
\usepackage{amssymb} 

\usepackage{mathptm}
\begin{document}

\title{Evidence Algorithm and\\ System for
Automated Deduction:\\
A Retrospective View\\
\small \;\; (In honor of 40 years of the EA announcement)}

\author{Alexander Lyaletski\inst{1} 
\and 
Konstantin Verchinine\inst{2}
}
\institute{%
\small Faculty of Cybernetics,
Kiev National Taras Shevchenko University\\
2, Glushkov avenue, 03680 Kiev, Ukraine\\
\small email:~{lav@unicyb.kiev.ua}
\and
\small Math-Info Department, Paris 12 University\\
61, avenue du General De Gaulle, 94010 Creteil, France\\
\small e-mail:~{verko@logique.jussieu.fr}
}
\date{}
\maketitle

\begin{abstract}
A research project aimed at the development of an automated theorem proving
system was started in Kiev (Ukraine) in early 1960s. The mastermind of the
project, Academician V.Glushkov, baptized it ``Evidence Algorithm'', EA%
\footnote{Below, we explain why this title was chosen; 
it was used first in \cite{Glu70}.}. 
The work on the project lasted, off and on, more than 40 years. 
In the framework of the project, the Russian and English versions of the 
System for Automated Deduction, SAD, were constructed. They may 
be already seen as  powerful theorem-proving assistants. 
The paper%
\footnote{The final publication of this paper is
available at www.springerlink.com} 
gives a retrospective view to the whole history of the
development of the EA and SAD. 
Theoretical and practical
results obtained on the long way are systematized. 
No comparison with similar projects is made.
\end{abstract}

\section{Introduction}

The research project entitled ``Evidence Algorithm''
was initiated by V.Glushkov
in the early 60-s in Kiev. At that time, some fundamental facts concerning 
formal proof search and opportunities (potential in most cases) to use
computers to find a proof, were already known. The domain that was called
``automated theorem proving''  (ATP or ``machine reasoning'' in the AI 
community) became a challenging one for logicians as well as for computer 
scientists (see
e.g. \cite{MD} for short history). There were hopes! Recall the title of
an early Hao Wang's paper: ``Towards mechanical mathematics'' \cite{wang60}.

V.Glushkov as he personally told us, was motivated by two main reasons: 

(1) To get an aid while verifying long and routine algebraic transformation
(as a working mathematician he obtained valuable results concerning  Hilbert's 
5th problem). 

(2) To try the strength of the existent computers pushing them to run on the
limits of their abilities. 

V.Glushkov formulated the main question in a slightly unusual way. 

Let us
consider some relatively well formalized mathematical theory, e.g. Lie
algebras. There are a small number of basic facts (axioms) which are
considered to be evident even for beginners. Let's apply simple purely logical
tools to obtain several consequences. They are also evident. Then one can
apply the same logical tools to the conclusions and so on. Are the results
still evident? If the conclusions were obtained by a programmed inference
engine, the answer is ``yes, they are''. From the viewpoint of this engine. But
probably not from the human point of view. Thus, provided the above-mentioned
engine, we would be able to prove/verify something that is not evident for
humans. Further to that, this ``evidence maintaining engine'' may be reinforced
with heuristics, proof methods, lemma application, definition expansion, and so
on. In this way, we could enlarge the notion of ``being evident'' to the extent
that might include nontrivial facts/theorems. Well, ``now do it, guys!''. 

That is why the algorithmic part of the project (and afterwards the project as
the whole) has got the name ``Evidence Algorithm'', 
EA, or $\exists\forall$ for fun. 

It was also already clear at that time that nobody would like to formalize the
mathematical knowledge/reasoning in the usual first order language. Hence a
formal but human-friendly language had to be developed to provide a 
possibility for 
the construction of a mathematical assistant system 
convenient for a wide range of scientists. 

So, three major components of such a system should be: 

(1) a powerful input language that must be close to the natural mathematical
language and easy to use;

(2) an inference engine that implements the basic level of evidence
(sometimes, we call it a ``prover'' below);

(3) an extensible collection of tools that reinforce the basic engine
(sometimes, we call it a ``reasoner'' below). 

In what follows, we give a short description in chronological order of what
has been done in each of the above-mentioned directions. 

Please note that our main goal
is to trace the long path of the project development and to recall the results
obtained. That is why the reference list is so long. For the same reason,
we could neither make any comparison with similar existing systems, nor give
an illustrative set of examples. Sorry for that. We frequently got an
impression that the automated reasoning community is not sufficiently 
acquainted with EA project
(for instance, SAD was not mentioned in F.Wiedijk's book 
\cite{FW}) though we think that
some ideas and results might be useful to know. 
We hope that the text given below will partially meet the lack of
such information. 

Note on the bibliography. Almost all papers published before 1992 were written
in Russian and therefore are hardly available now. We translated the titles
and put them onto the list just to indicate what was done in the old time.
All the papers are listed in chronological order.

The rest of the paper contains three parts according to the three periods in
the history of the EA project. They are as follows. 

The first one: 1962 - 1970. We call it ``Pre-EA Stage'' below.

The second one: 1970 - 1992. It is called ``EA and Russian SAD'' below. 

The third one: 1998 - nowadays. Below it is called 
``Post-EA Stage and English SAD''. 

Several final remarks conclude the paper.

\section{Pre-EA Stage (1962--1970)}

Few people remember now that the Soviet computer history began in
Kiev. The first von Neuman computer was assembled and tested at the turn of
1950 in a small laboratory headed by the academician S.Lebedev. In 1955
S.Lebedev left for Moscow and the director of Kiev Institute for Mathematics,
prominent mathematician B.Gnedenko invited V.Glushkov to take the
supervision of the laboratory (which was transformed into the Institute of
Cybernetics 5 years later). 

On the other hand, at that time there was a powerful logic, linguistic and 
algebraic team at the mathematical department of the Kiev University. 
Professor L.Kaluzhnin who was the head and the heart of the team, invited 
V.Glushkov to join their efforts. 

So the Kiev school in the ATP domain appeared really at the borderline
of computer science and mathematical logic. 

In 1962, V.Glushkov published a paper \cite{Glu62eng} where he analyzed several
rather simple proofs in Group Theory and suggested that the proofs might
be built automatically with the help of a not too complex procedure. The idea
attracted three people who began their research on the subject:
A.Letichevsky, one of the first Glushkov's disciples, (in 1962), F.Anufriev
(in 1962) and V.Fedyurko (in 1963). A bit later, V.Kostyrko and Z.Aselderov
had joined the team. The first-time approach to the problem was purely 
empirical -- they analyzed a lot of proofs taken from textbooks, 
monographs and articles for
trying to formalize all them and to find (almost by  feeling) 
methods, heuristics and representation details that might help 
to construct a proof of a theorem under consideration automatically. 
As a result an algorithm of proof search in Group Theory was
constructed and even implemented (the corresponding program run on 
the monstrous Ural-1 computer). 
The first communication about it was done at the First
All-Union Symposium on the Machine Methods of Logical Inference Search, that
took place in  Lithuania in 1964 \cite{M64MS} (see also \cite{L86MP}). Later
a paper on the subject was published \cite{AnuFedLetAseDid66} (and translated
afterwards into English). 

The algorithm, though being comparatively simple, contained nevertheless: 

- a method of inference search for some class of first-order formulae;
 
- a reduction technique for simplifying search  space;

- a collection of heuristics (e.g. the inclusion relation was exploited);

- special methods of equation solving. 

So we can say that it was the first problem-oriented prover for Group
Theory. Here is an example of proved theorem: ``The centralizer $Z$ of any
subgroup $P$ is a normal subgroup of the normalizer $N$ of $P$''. 

The above-mentioned proof-search method resembled, in some sense, well known
backward chaining, but some features were added to make it applicable to
non-Horn formulae. Later on, the method was generalized by F.Anufriev and
extended to the whole first-order classical logic without equality
\cite{Anu67eng,Anu69eng}. It can be interpreted as a goal-oriented
sequent calculus not requiring skolemization and using an analog of
Kanger's notion of substitution admissibility. 
Later, the method was transformed into a correct
and complete sequent calculus \cite{Mala74aeng} with skolemization. 
It had got the name
``Auxiliary-Goals Search calculus'' (``AGS calculus'' below) 
and served as a prototype for various
sequent-type inference engines of the EA project. 

Solving equations in free groups became the subject of Z.Aselderov's 
PhD thesis \cite{Ase68eng}, which was defended in 1968. 

Now let's cast a glance at the list of required components of the conceived EA
system. No convenient input language was yet proposed at the time being. On the
other hand, it was difficult to continue the project without it. To see why,
try to convert the theorem above to the first-order language. 
On this subject, there were only
two Kaluzhnin's papers:  \cite{Kal62eng} and
\cite{KalKor64eng}. Some time later, 
V.Kostyrko made an attempt to solve the problem and after some
period, a paper was published \cite{GluKosLetAnuAse70eng} where a contour of
such a language was outlined. The main idea was as follows. 

Let's consider an atomic first-order formula. It is always of the form
$R(t_1,t_2,\dots ,t_n)$ where $R$ is an $n$-ary relation symbol, whereas a
``natural'' atomic statement is of the form 
$<subject\; group>\; <predicate\;group>$. Well, one can: 

 - select an argument among $t_1,t_2,\dots ,t_n$ , say $t_1$, 

 - consider it as the subject, 

 - ``reduce'' $R$ to something $(n-1)$-ary $N(t_2,\dots ,t_n)$,

 - add a new connector to ``attach'' $t_1$ to $N(t_2,\dots ,t_n)$ ($\epsilon$
   was chosen in the original version).  

Now $R(t_1,t_2,\dots ,t_n)$ can be written as $t_1$ 
$\epsilon$ $N(t_2,\dots ,t_n)$
and read as $t_1$ {\em is a} $N(t_2,\dots ,t_n)$. For example, $Subgroup(H,G)$
gives $H$ $\epsilon$ $Subgroup\_of$ $G$ and so on. 
Was it not more than syntactic ``sugar'' or one could gain 
something interesting with it? 
Below we demonstrate what was made in this direction later. 

For the completeness of the description of that time, 
it may be needed to remind the
last implementation of the propositional part of Anufriev's procedure, 
which was made by A. Malashonok on the BESM-2 computer at the 
beginning of 1970 \cite{AnuKosMal72}.

\section{EA and Russian SAD (1971 - 1992)}

In 1970, V.Glushkov published one more paper on the subject \cite{Glu70}. 
At that time, he associated the progress in the domain of ATP 
with the general tendency to make
computers more intelligent (see also \cite{GluKap72}). As to the project in
question (except the fact that it had 
got its name ``Evidence Algorithm''), 
V.Glushkov  emphasized the importance of a
natural formal language for writing mathematical texts in \cite{Glu70}. 
We should also note
that for the first time, the term ``automated theorem proving'' was used instead
of ``automatic theorem proving'' and the problem of how to construct 
something like
an ``interactive proof environment'' was explicitly formulated. In fact, a proof
assistant was conceived at that time.  

It seems that somewhen in the middle of 1970, V.Glushkov decided to add ``young
forces'' to the existing EA team, and he charged one of his former pupil
V.Bodnarchuk to became the leader of the new team. 
At that time, V.Bodnarchuk was the head of a computer 
department; its members had just finished their work under a
specialized mini-computer for engineering computation with the language 
``Analitik'' \cite{GluBod70eng}.

The input language ``Analitik'', being convenient for engineers and 
having its hardware implementation, was one of   
the distinguishing features of the computer, and V.Bodnarchuk 
was its main creator. Besides, V.Bodnarchuk was very intimate with
L.Kaluzhnin, and these exerted great influence on the 
development of the EA.

At the same time, four young people became the postgraduate students at the
Institute of Cybernetics, both authors were among them. 

Two of them, A.Degtyarev and K.Verchinin, were graduated from the Mechanical and 
Mathematical Faculty of the Moscow State University. Other two, A.Lyaletski and
N.Malevanyi -- from the Cybernetics Faculty of the Kiev State University. So,
we had joined the EA team and V.Bodnarchuk became our ``local supervisor''
(the global one was V.Glushkov). We were young, full of energy and illusions... 

At the very beginning, V.Bodnarchuk has formulated the following tasks: 

- careful revision of everything that was done previously by the ``old team''; 
 
- detailed analysis of mathematical texts in various domains; 

- preparation of two surveys: 
           (1) of combinatorial proof-search methods (published, see 
\cite{KapKosLyaDegMalAnuAse72}) and
           (2) of using heuristics in proof search (published, see 
\cite{KapDegLya73})

The revision of the existing version of the AGS method 
demonstrated that, first, 
the use of the Kanger's notion of substitution admissibility instead 
of skolemization complicates drastically an eventual
implementation and, second, the method requires a special technique 
for equality handling. So, to advance the whole project, one needed: 

 - either to improve the AGS method paying special attention to redundancy
   avoidance and equality handling, or to adapt one of existing combinatorial 
methods
   of proof search for the role of inference engine in the EA project; 

 - to develop a practically usable version of the ``mathematical language''
   along with the whole syntactical service around it; 

 - to find a convenient formalization of what is frequently used in
   mathematical texts to make them available for human reader -- ``proof
   method'', ``proof scheme'', ``lemma application'', 
``definition dependency'',
   etc. 

- to find methods of what is called ``knowledge management'' now, e.g. 
to try to
   understand what the ``relevancy relation'' on mathematical facts might be; 
 
 - to develop an implementation base (it became clear at the very beginning
   that experimental work was strongly needed and it could not be done in 
the paper-and-pencil mode).

We began in quite favorable setting. Two circumstances should be especially
noted. 
At that time, it was easy to establish scientific contacts in the
ex-USSR and we have done that: with famous Leningrad logic school, with
excellent Novosibirsk logic school (founded by A.Maltsev), with strong Moscow
logic school, with linguists, psychologists, etc. The second point is that
last-year students of the new Cybernetics department of the 
Kiev University used to pass their six month professional training 
at the Institute of Cybernetics. In this way
the second EA team had got two very capable young researchers: 
A.Zhezherun (in 1973) and M.Morokhovets (in 1978).  


\subsection{Theoretical work}

Here is a brief description of research interests and results obtained by
members of the second EA team. 

At the beginning, A.Degtyarev studied the role of heuristics in formal
proofs. He restricted himself with linear algebra and showed that for large
class of theorems, the proof search (by resolution with paramodulation) may be
controlled in a way and reduced to the problem of finding solution to a set
of linear equations \cite{Deg73,Deg74aeng}. It was quite interesting result
but A.Degtyarev did not continue that direction and devoted himself to the
problem of equality handling in resolution-like methods. 
As a ``side effect'' he
obtained an efficient unification algorithm (published later in
\cite{Deg75eng,Deg80eng}) that was based on the same principles that the
well-known Martelli and Montanary algorithm \cite{MM82UA} formulated
later. 

His main results concern various paramodulation strategies and the
problem of compatibility the paramodulation rule with term orderings. The most
known is so called monotonic paramodulation \cite{Deg79eng,D79SP,Deg82aeng}
subsequently used in many other researchs on the subject.  

A.Lyaletski occupied himself with the careful analysis of combinatorial proof
search methods trying to put them in a common setting and find (or build) the
best candidate for a resolution-type  
inference engine. He suggested a modification of
the resolution rule which operated with more general objects than clauses --
conjunctive clauses or c-clauses. (Later, V.Lifschitz \cite{L89IM}
independently proposed something similar and called them ``super-clauses''). 
Two different c-clause calculi were build \cite{Lya75eng,LyaMal75eng} which
permitted to reformulate well-known Maslov's Inverse Method \cite{M64IM} in a
resolution-like manner. 

Another problem was the skolemization. Is it bad or not? 
Anufriev's method did not use skolemization, but it adds new entities
as in the case of Kanger's method. On the
other hand, skolemization simplifies the algorithmic part of proof search
methods. A.Lyaletski found an original notion of admissible substitution that
allowed him to get in some sense a compromise. He built a series of sequent
calculi with resolution style inference rules, that, on one hand, don't require
skolemization and, on the other hand, are not less efficient than the usual
resolution calculus (\cite{Lya79aeng,Lya79beng,Lya82a}).

K.Verchinine was strongly involved in the language problem. We had to
formalize mathematical texts, not only isolated statements. A text may be
considered as a structured collection of sections: chapters, paragraphs,
definitions, theorems, proofs, etc. So a part of the language was designed to
represent this structure, its ``semantics'' was given by the ``trip rules''.
Another part served to formalize a statement. New units were added to
the standard first-order syntax which permitted to use nouns, adjectives,
special quantifiers, etc. The language was developed and has got the name TL
-- Theory Language \cite{GluKapLetVerMal72,GluVerKapLetMalKos74eng}. Here is a
formal TL phrase: ``there is no remedy against all deseases but there is a
desease against all remedies''. 
(That time the vocabulary as well as the syntax
was certainly Russian.) 

Two kind of semantics were defined for that
part: a transformational one (an algorithm to convert a TL statement into its
first-order image) and another one -- in the traditional set-theoretical style
where $\epsilon$ was interpreted as the membership relation \cite{Ver73}. The
last semantics permitted to define the ``extension'' of every notion (e.g. the
extension of ``subgroup$\_$of G'' is the class of all subgroups of G) and to
introduce a structure on the set of notions which restrict quantifiers in the
given sentence. That structure was called ``situation'' and was used in attempts
to formalize a relevancy relation. 

At the beginning, A.Zhezherun took active part in the TL language
development. He designed and implemented the whole syntactic service for the 
linguistic part of the future system. As usual, there were funny side
effects of the work. For instance, computer linguists have always searched
for some invariant (called profound semantic structure) that could be used in 
machine translation algorithms. A.Zhezherun and K.Verchinine showed that the
first-order image of a TL statement can play the role of such invariant. So
just changing the superficial decorations in some regular way, one can
translate mathematical statements from Russian into English and vice versa
(provided the dictionary). A.Zhezherun wrote a program to play with, and it
worked surprisingly well! Besides, he studied the opportunity to formalize
mathematical reasoning in a higher-order logic and proved in particular the
decidability of the second-order monadic unification \cite{Zhe79}. 

M.Morokhovets occupied herself with the problem of ``reasoner'' (see
above). As the reasoner must have a prover to cooperate with, the last was
badly needed. The AGS based prover didn't fit well to that purpose, so we
decided to develop and implement a resolution-and-paramodulation  based prover
with a flexible architecture that could be adapted to various strategies and
auxiliary inference rules. M.Morokhovets has done it. The first observation
showed that some particular premises are strongly responsible for the search
space explosion. The transitivity axiom clearly is among them. M.Morokhovets
proved that for some large class of transitive relations, this axiom may be
eliminated and replaced by a special inference rule which can be controlled to
shorten the search space \cite{Mor85}. 


Another idea was to use the fact that all quantifiers in the TL statement are
restricted (bounded). Is it possible to ``forget'' the restrictions, to find an
inference and then just to verify that all substitutions are correct 
w.r.t. these restrictions (bounds). 
M.Morokhovets has found several classes of statements for
which the answer is ``yes'', and has implemented corresponding procedure
\cite{VerMor83b}. One more question was as follows. Let's suppose that a
conjecture is proved and the resolution style inference is constructed. How to
present it in a human readable form? The set of conversion rules that permit
to do it (based on an early result of K.Verchinine), was designed and
implemented by M.Morokhovets, too. 

\subsection{Experimental work}

Certainly, some computer experiments have been done from the very beginning of
EA project development (it was one of Glushkov's ideas -- to be permanently
accompanied with computers while doing theoretical research). Still in 1971
K.Verchinine used the syntactic tools taken from another system (developed in
the same department) to implement a part of TL grammar. 
A.Malashonok have programmed AGS prover to make local experiments with. Also
local experiments with paramodulation strategies were maid by
A.Degtyarev. N.Malevanyi began to prepare something like a specialized library
for future experiments on the BESM-6 machine -- another Lebedev's creation --
one of the most powerful computer in the ex-USSR. 

Systematic programming was initiated after 
A.Zhezherun appeared. He became the main
designer and programmer of the system for mathematical text processing. But no
doubt, we all were involved in programming. At that time, the IBM System 360/370
(cached under the name ``ES Line Computer'') was admitted in the ex-USSR as the
main platform. With the native operating system and the PL/1 as the main
programming language -- what a hell!!!  

Nevertheless the work advanced and the first experiments with the whole system
were done in 1976/1977. The main task was formulated as mathematical text
verification and may be presented as follows. 

Let a TL text be given. The system can: 

 - parse the text informing the user about syntactic errors (if any); 

 - convert the text to some tree-like internal form; 

 - run the main loop: choose a goal sentence to verify and find its logical
   predecessors;  

 - construct an initial proof environment for one of available provers%
\footnote{At that time, the SAD system prover was constructed 
and implemented on the base 
of an original sequent-type calculus \cite{DegLya81eng}. 
It had the following features: it was goal-oriented, 
skolemization was not obligatory, and equality handling was separated
from deduction. Now, the native prover of the current (English) SAD 
possesses the same features.};

 - start the prover and wait; 

 - if the prover fails then ask to help;

 - if the prover succeeds then output the proof, choose the next goal and
   repeat the main loop until the end of the text be reached. 

The first public presentation of the system in question was made at the
All-Union symposium ``Artificial intelligence and automated research in
Mathematics'' (Kiev, Ukraine, 28-30 November 1978). It worked! 

In 1980, V.M. Glushkov gave the name ``System for Automated Deduction''
(SAD) to the implemented system and it has this name now. 

The further work consisted in improving the system and adding new features to
it. We extended the mathematical texts library and developed a conception of
further extension of TL language with ``imperative'' (algorithmic)
constructions. A method of using auxiliary statements in proof search
(based on the notion of situation) was implemented by V.Atayan
\cite{Ata81eng}. Efficient paramodulation strategies were added and tested by
A.Degtyarev. A resolution-based prover was implemented by M.Morokhovets.

In the meantime four PhD thesis were defended at the Institut for Cybernetics:
A.Zhezherun has got his PhD in 1980 \cite{Zhe80aeng}, A.Lyaletski
\cite{Lya82a}, A.Degtyarev \cite{Deg82beng} and K.Verchinine \cite{Ver82eng}
-- in 1982. M.Morokhovets' thesis was in preparation.  

We understood that to advance the project we need to try the SAD system in
some more or less practical applications. One possible application was the
automated program synthesis and we established a contact with professor Enn
Tyugu (Tallinn, Estonia) and his team. Another interesting application was the
deduction tool for expert systems. The problem is that 
classical logic is rarely used in this domain. So, the question appeared: is it 
possible to
adapt SAD for the inference search problem in non-classical logics?%
\footnote{Later, a theoretical answer on this question was obtained 
in a number of papers of A.Lyaletski (see, for example,
\cite{KL06IIS,L08SYNASC}); from this point of view, 
some researches on Herbrand theorems  
(\cite{L06AMAI,LK06JELIA,LL06LC,L09SYNASC}) 
also may seem to be interesting.}.

But everithing comes to its end. Sooner or later.

\subsection{Team evolution (or the sad part of the SAD history)}

Already in the end of 1972, V.Bodnarchuk falled seriously ill and, actually, 
he abandoned the research activity for a long time. From 1973 to 1975
F.~Anufriev, Z.~Aselderov, V.~Kostyrko, and A.~Malashonok left the team
because of various reasons, they never came back to the subject area
afterwards. In 1982, V.Glushkov was dead. The administration style in the
Institut for Cybernetics changed and we were not the favorit director's team
any more. In the middle of 1983, A.~Lyaletski and A.~Zhezherun left for the
Kiev University. In 1984, K.~Verchinine moved to another department and changed
his research area. Finally, in 1987, A.~Degtyarev left for the Kiev
University, too. M.Morochovets stayed at our former department of the
Institute of Cybernetics. The EA team did no more exist...

\section{ Post-EA Stage and English SAD (1998-nowadays)}

In 1998, the Evidence Algorithm project moved into a new stage.
That year the INTAS project 96-0760 ``Rewriting techniques and
efficient theorem proving'' started and brought financial support
for resumption of work on SAD. The new working group included Alexander
Lyaletski at Kiev National University (KNU), Marina Morokhovets at the
Institute of Cybernetics in Kiev, Konstantin Verchinine at Paris 12 University
in France, and Andrei Paskevich, fourth-year undergraduate student of KNU.

The work started in 1999, with re-implementation of the TL language on IBM
PC. The programs were written in C on the Linux platform. In a year, towards
March 2000, parsing and translation of TL sentences into a first-order
language was implemented. The English-based version of TL had been given
the name ForTheL, an acronym for ``FORmal THEory Language'' 
(also a Russian word
meaning ``trick'' or ``stunt''). 
The language was presented firstly at
the Fifth International Conference ``Information Theories and Applications'' 
in September 2000 in Varna, Bulgaria \cite{PV00ITA}.

The same summer the work started on re-implementation of the deductive tools
of SAD. By January 2001, A.Paskevich created the first prototype of the prover
(the prover had gotten the name ``Moses''). 
A bit later the technique of admissible
substitutions by A.~Lyaletski  which permitted to dispense with skolemization 
and preserve the initial
signature of a proof task, was also implemented. Later, the equality
elimination procedure by Brand \cite{Bra75} was added to handle the problems
with equality. By June 2001, the complete ``workflow'' of the initial SAD: from
ForTheL text to first-order representation to proof task to proof tree, was
reestablished. Of course, a lot of functionality of the previous
implementation has not been transferred into the new system.

In September 2001, A.~Paskevich started his doctoral study under
the joint supervision of Konstantin Verchinine and Alexander Lyaletski.
His work aimed at the development of a new, two-level architecture of
a mathematical assistant.

In the first prototype of the English SAD system, the reasoner was virtually
non-existent. The theoretical development of the reasoner started with the
work on ``local validity'', which allowed to perform sound logical inferences
inside a formula, possibly under quantifiers. This technique could provide a
basis for in-place transformations (such as definition expansions) as well
as for handling of partial functions and predicates \cite{P02CALC}.

By the end of 2003, tools for supporting proofs by case analysis and
by general induction (with respect to some well-found ordering) were
implemented in the  SAD. In 2004, an experimental support for binding
constructions, such as summation and limit, was also added \cite{LPV06JAL}.   

An algorithm for generation of atomic local lemmas
was constructed and implemented: these lemmas help to prove a lot of simple
statements without using a prover at all. 

An interesting feature of the SAD is that the prover does not depend on the
rest of the system. It means that various provers can be used as the system
inference engine (provided the interface be written). The following ones 
were used
in our experiments: SPASS \cite{SPASS}, Otter \cite{Otter}, E Prover
\cite{Eprover}, Vampire \cite{Vamp} and Prover9 \cite{prover9}.

In July, 2007, the ``enriched'' SAD system was presented at the 21st Conference
on Automated Deduction in Bremen, Germany \cite{VLP07CADE}. A.~Paskevich has
made several improvements since then. The current version of the system is
freely available at  http://www.nevidal.org . Here is a short list of texts
(proofs) that were successfully verified by the SAD: 
Tarski's Fixed Point theorem,
Newman's lemma,
Chinese Remainder theorem,
Infinite Ramsey theorem,
``The square root of a prime number is not rational'',
Cauchy-Bouniakowsky-Schwartz inequality for real vectors,
Fuerstenberg's proof of the infinitude of primes.

Finally, note that the EA project leaded to 
the carrying out of new investigations in automated reasoning
(see the last publications in the reference list).

\section{Conclusion}

Let's imagine an ideal Mathematical Assistant. What its architecture might be
from the EA position? 

A user communicates with the system with the help of texts written in a
high-level formal input language close to the natural one. She or he submits a
problem like ``verify whether the given text is correct'' or ``how to prove
the following statement'', or ``what is the given text about'' and so on. The
text, provided being syntactically correct, is treated by the part of the
system that we call  ``reasoner''. The reasoner analyzes the problem and
formulates a series of tasks that it submits to the inference engine, a
prover. If the prover succeeds, the resulting conclusion (e.g.~human-readable
proof) is given to the user and the game is over. If it fails then a kind of
``morbid anatomist'' makes a diagnosis and supplies it to the reasoner who
tries to repair the situation. In particular, the reasoner can decide that an
auxiliary statement (lemma) might be useful and start the search for those in
the mathematical archives. To do that it submits a request to the archive 
service, we call it ``librarian''. After getting an answer, the reasoner 
begins a new proof search cycle with the modified problem and 
the process goes on.

The user can interact with the system by playing for the reasoner, 
librarian, for the morbid anatomist (provided that she or he understands 
the internal prover's life) or for the prover itself, deciding whether 
a given conjecture should be considered as valid.

Where we are with respect to the ideal? Optimistic answers are welcome.

\section{Acknowledgements}

We are grateful to our teachers. We are grateful to everybody who worked side
by side with us during all that time. Special thank to the referees for their
patience.


\end{document}